\documentclass[letterpaper, 10 pt, conference]{ieeeconf}  

\IEEEoverridecommandlockouts                              

\usepackage[T1]{fontenc}
\usepackage{ulem}
\usepackage{graphicx}
\usepackage{float} 
\usepackage{url} 
\usepackage{color}
\usepackage{xcolor}
\usepackage{amsmath}
\usepackage{mathtools}
\usepackage{hyperref}
\hypersetup{pdfborder={0 0 0}}
\usepackage[table]{xcolor}
\usepackage{booktabs} 
\usepackage{makecell}

\newcommand{\qasep}{\newline\rule{0em}{0.5\baselineskip}}

\title{\LARGE \bf An Integrated System for WEEE Sorting Employing X-ray Imaging, AI-based Object Detection and Segmentation, and Delta Robot Manipulation}

\author{Panagiotis Giannikos$^{1}$, Lampis Papakostas$^{1}$, Evangelos Katralis$^{1}$, Panagiotis Mavridis$^{1}$, \\ George Chryssinas$^{1}$, Myrto Inglezou$^{2}$, Nikolaos Panagopoulos$^{1}$, Antonis Porichis$^{2}$,\\ Athanasios Mastrogeorgiou$^{1,3}$, Panagiotis Chatzakos$^{2}$
\thanks{$^{1}$P. Giannikos, L. Papakostas, P. Mavridis, E. Katralis, G. Chryssinas, N. Panagopoulos \& A. Mastrogeorgiou are with THL (Tech Hive Labs), \{panagiotis.giannikos, lampis.papakostas, panagiotis.mavridis, evangelos.katralis, 
georgios.chrysinas, nikolaos panagopoulos, athanasios.mastrogeorgiou\}@thlabs.eu, 280 Kifisias Ave., 152 32 Halandri, GR.}%
\thanks{$^{2}$Panagiotis Chatzakos, Myrto Inglezou and Antonis Porichis are with the AI Innovation Centre, University of Essex, \{p.chatzakos, m.inglezou, a.porichis\}@essex.ai, Wivenhoe Park, Colchester CO4 3SQ, UK.}%
\thanks{$^{3}$Athanasios Mastrogeorgiou is with the Control Systems Lab, School of Mechanical Engineering, NTUA, GR. (email: amast@central.ntua.gr)}%
\thanks{$^{}$ This work has received funding from the European Union’s Horizon 2021 research and innovation program under grant agreement N° 101070321. The authors thank Varex Imaging for providing the X-Ray imaging detectors.}%
}

\begin{document}

\maketitle
\thispagestyle{empty}
\pagestyle{empty}

\begin{abstract}
Battery recycling is becoming increasingly critical due to the rapid growth in battery usage and the limited availability of natural resources. Moreover, as battery energy densities continue to rise, improper handling during recycling poses significant safety hazards, including potential fires at recycling facilities. Numerous systems have been proposed for battery detection and removal from WEEE recycling lines, including X-ray and RGB-based visual inspection methods, typically driven by AI-powered object detection models (e.g., Mask R-CNN, YOLO, ResNets). Despite advances in optimizing detection techniques and model modifications, a fully autonomous solution capable of accurately identifying and sorting batteries across diverse WEEEs types has yet to be realized. In response to these challenges, we present our novel approach which integrates a specialized X-ray transmission dual energy imaging subsystem with advanced pre-processing algorithms, enabling high-contrast image reconstruction for effective differentiation of dense and thin materials in WEEE. Devices move along a conveyor belt through a high-resolution X-ray imaging system, where YOLO and U-Net models precisely detect and segment battery-containing items. An intelligent tracking and position estimation algorithm then guides a Delta robot equipped with a suction gripper to selectively extract and properly discard the targeted devices. The approach is validated in a photorealistic simulation environment developed in NVIDIA Isaac Sim and on the real setup.

\end{abstract}
 

\section{Introduction}

As the volume of Waste from Electrical and Electronic Equipment (WEEE) continues to grow rapidly, the need for efficient collection and recycling processes becomes more and more critical \cite{Globalewastemonitor2024, Assessingtherisk2024}.
Batteries, which are often found in WEEE, contain a significant amount of hazardous materials that, if not managed accurately at the end of their life cycle, can pose serious threats to human health, recycling facilities and the environment \cite{Formalrecyclingofewaste2014, Efficiencyandfeasibility2008}. Moreover, advances in battery technology, particularly in power density and compactness, as seen in modern lithium-ion batteries, have amplified the risk associated with their recycling, including fire hazards and explosions \cite{WEEEfires2020, Lifecycleassessment2017}. 
Currently, battery removal from WEEE is primarily carried out manually, due to the high complexity and diversity of electronic devices, exposing workers to potential safety hazards \cite{Challengesandbestpractices2024}.
At the same time, proper battery separation is essential, as batteries contain valuable metals that can be recovered, thus mitigating the environmental impact of raw material extraction \cite{Ewasterecycling2023}. 

\begin{figure}[!ht]
  \centering
  \includegraphics[width=0.48\textwidth]{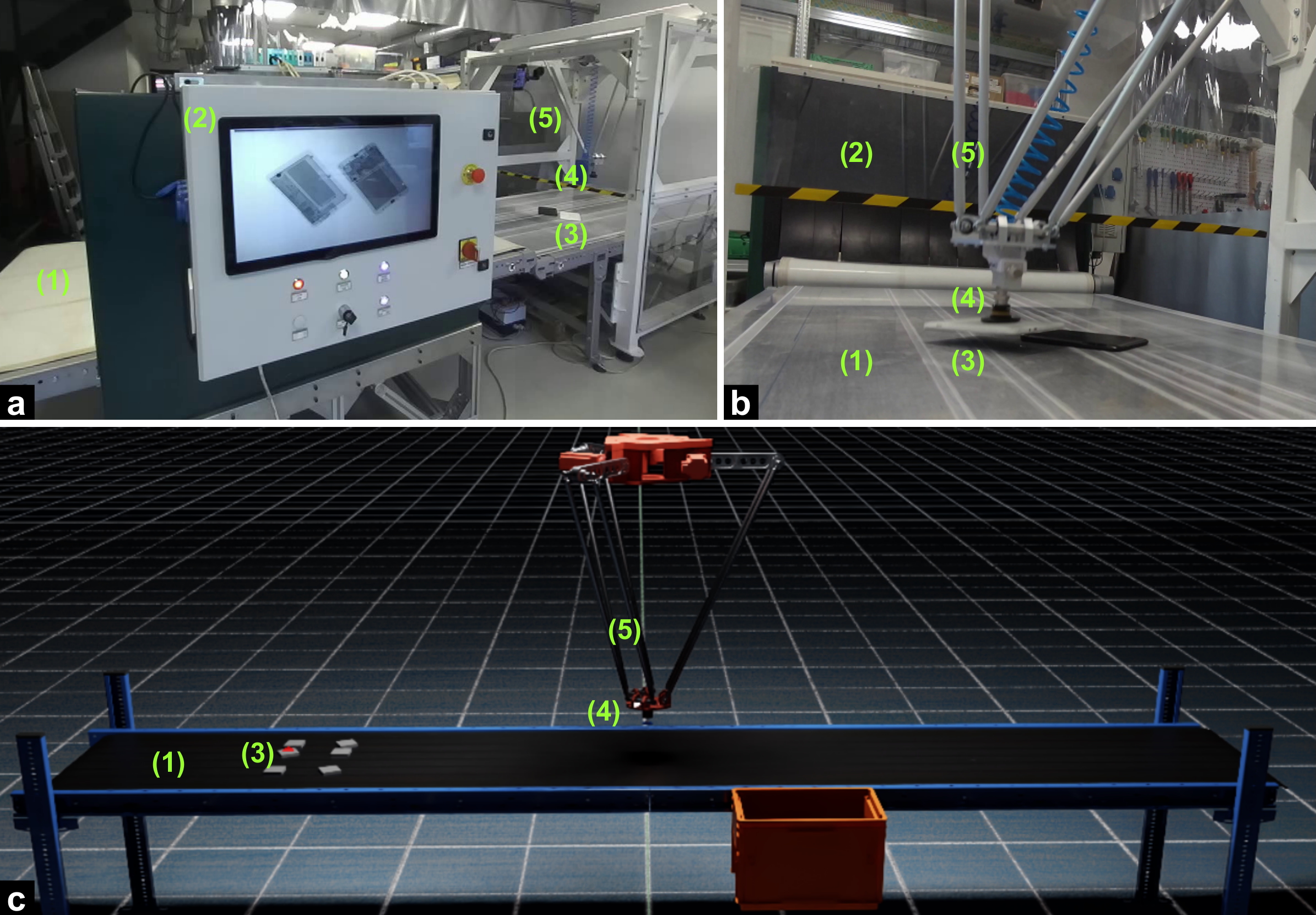} 
  \caption{The developed WEEE sorting system. (1) The conveyor belt, (2) the X-ray transmission imaging system, (3) the WEEE devices and (4) the suction gripper attached to (5) a delta robot. a) X-ray imaging module, b) Delta robot, c) System in the NVIDIA Isaac Sim simulation environment.}
  \label{fig: Grinner setup}
\end{figure}


A plethora of systems have been proposed to inspect, classify or dismantle WEEE. In \cite{Inlinesortingsystemwithbatterydetection2024, Enhancingbatterydetection2024, X-raybasedroboticewastefractionation2022, Detectionandrecognitionofbatteries2021} interior components of WEEE are examined utilizing X-ray transmission imaging systems but with single energy spectrum acquisition, which has moderate discriminative capabilities between thin and dense materials. The study in \cite{Designofaroboticsystemforbattery2022} proposes a black-box approach to tablet dismantling, focusing only on the overall presentation of the procedure. In \cite{Roboticsortingofbatteriesusingvisualfewshot2023, Roboticsortingofusedbuttoncellbatteries2018, Sortingsystemforewasterecycling2019}, vision based methods are presented which specialize in the sorting of different types of batteries, different chemistry of button cell batteries and components of a PC motherboard respectively, without being able to generalize to a variety of WEEE. In the aforementioned papers, as well as in \cite{BatterydetectionofX-rayimages2022}, a deep learning model, e.g. Mask R-CNN \cite{MaskRCNN2018} and YOLO \cite{YOLO2016}, is employed to perform battery detection on acquired images and output detailed instructions to a human supervisor for removing the batteries. However, no fully autonomous sorting system has been developed so that it can easily detect batteries in an assortment of WEEE and remove the respective devices from the recycling line.


Our work presents the integration of a specialized dual-energy X-ray transmission imaging subsystem with advanced preprocessing software, enabling high-contrast image reconstruction capable of effectively differentiating between dense and thin scanned materials. The proposed approach provides a comprehensive solution specifically designed for using a delta robot to sort flat electronic devices containing batteries, which constitute a significant fraction of hazardous WEEE \cite{Globalewastemonitor2024, WEEEfires2020}.
Figure  \ref{fig: Grinner setup} illustrates the entire robotic sorting pipeline. Initially, WEEE items are transported via a conveyor belt through an intelligent, high-resolution X-ray imaging system designed to detect battery-containing devices. Accurate battery detection and device segmentation are performed using YOLO \cite{YOLO2016} and U-Net \cite{UNet} neural network models, respectively. Subsequently, based on model predictions from the X-ray images, a tracking and future-position estimation algorithm determines the target object’s location. Finally, a customized delta robot equipped with a suction gripper extracts identified battery-containing WEEE from the conveyor line and discards it appropriately. Additionally, an Isaac-based simulation environment is integrated into the framework to evaluate system performance and generate optimized robot trajectories, ensuring precise, reliable, and efficient sorting operations.

This paper is organized as follows: Sect. II describes the system overview, covering the X-ray transmission imaging, the AI-based battery detection, WEEE segmentation and future position estimation software as well as the robotic handling of the WEEE. Sect. III outlines the ISAAC simulation of the WEEE sorting system, including the delta robot modeling, kinematics and trajectory planning. Following that, Sect. IV describes the sorting procedure implemented and the software architecture of the system. Sect. V presents the experimental results and Sect. VI concludes with a discussion of future research directions.


\section{System Overview}

\subsection{X-ray Transmission Imaging}
\label{sec: X-ray transmission imaging}





When X-rays interact with matter, they undergo absorption, scattering, or transmission in a probabilistic manner that varies depending on their energy and the material's atomic number and density. Specifically, when an incident mono-energetic (single energy) photon beam with an intensity $I_0$ collides perpendicularly with an absorbing material with linear attenuation coefficient $\mu$ and thickness $x$, the intensity $I$ passing through the absorber can be evaluated by the Beer-Lambert law: \(I = I_0 e^{-\mu x}\).
Considering that X-rays are a form of high-energy electromagnetic radiation, they are able to penetrate a wide range of materials, but not dense ones. This exponential difference in the X-ray transmission, as expressed in the Beer-Lambert law, is the key to X-ray imaging, allowing the creation of detailed images of the internal composition of objects that are opaque to visible light and could not be inspected otherwise.

The developed X-ray transmission imaging system (Fig. \ref{fig: Grinner setup}a) consist of a conveyor belt by Dorner \cite{Dorner}, the IXS1620 X-ray source by VJX \cite{VJXIXS1620} and the DC-TDI800 X-ray line-scanner by Varex Imaging \cite{VareximagingDC-TDI}. A rotary encoder connected with an Arduino micro-controller is mounted on top of the conveyor, to measure and record its speed. The IXS1620's performance is characterized by a quick rise time and a maximum output power of $200$ W in continuous mode, which ensures the minimum radiation exposure, while emitting a large amount of high-energy X-rays. The intensity of the generated X-rays is sufficient to penetrate the examined WEEE items. The employed X-ray detector utilizes advanced direct-conversion photon-counting technology, ensuring exceptional sensitivity and enabling high-speed scanning operations. Activation of the X-ray line-scanner is precisely synchronized with the motion of objects on the conveyor belt via encoder feedback. As a result, the imaging system produces high-resolution X-ray images characterized by a superior contrast-to-noise ratio and minimal blur compared to conventional scintillator-based detectors (see Fig. \ref{fig: advanced X-ray imaging}a).. Additionally, the consistency of the conveyor belt velocity is confirmed by the accurate alignment and clarity of the X-ray images reconstructed from sequential scanned lines.

Another key aspect of the DC-TDI800 X-ray detector is that it supports native dual-energy imaging with spatial registration and energy separation. In this process, each incoming photon in every pixel is individually counted within one of two energy ranges, resulting in the generation of two simultaneous energy-domain images; High Energy (HE) by photons with energy \(E_\gamma > 60\) keV and Total Energy (TE) by photons with energy \(E_\gamma > 10\) keV. This dual-energy approach enhances material discrimination and composition analysis since it leverages the different absorption characteristics of materials at two distinct X-ray energy spectra. In the context of WEEE, this becomes an invaluable tool as it can be used to accurately identify and classify different types of batteries according to their chemical composition.

The X-Ray detector outputs data in the form of 16-bit integer arrays, which represent the photon counts for each pixel in the detector. The raw photon counts have to be normalized and adjusted for small discrepancies between pixels to ensure that a usable image is reconstructed. Each array returned by the detector corresponds to a line with a width of the same number of pixels as the pixels in the detector and a specific line height. These lines can be placed in a buffer and combined accordingly to create images. In our case, due to data bandwidth limitations, $4\times4$ pixel binning is applied, which reduces the image array size by a factor of $4$, this is crucial to ensure fast and efficient processing of the X-Rays through the entire detection and segmentation pipeline. The software that processes, captures, and passes the X-Rays to the detection and segmentation pipeline is custom built. The detector produces two arrays every time, each one corresponding to the different energy thresholds. Every time a new line is produced by the detector, it is placed in a buffer in such a way that the oldest line in the buffer is removed. The buffer length can be chosen to be arbitrarily large depending on the use case. This buffer exists for both energy levels and it corresponds to one image which is constantly updated. The produced images are passed through battery detection and device segmentation pipeline (see Section \ref{ai_detection}) in such a way that there is a $50\%$ overlap between images. This means that the bottom half of an image is the top half of the image that follows. This ensures that all objects are captured \& passed to the developed pipeline. The two arrays corresponding to each energy level are converted from $16$-bit to $8$-bit and reconstructed as an image where each channel corresponds to a different energy level.

\subsection{AI-Based Battery Detection and WEEE Segmentation} \label{ai_detection}

Once an X-ray scan is captured, the next step involves detecting the locations of all batteries present in the image and matching them to the corresponding devices on the conveyor. To achieve this, we employ two distinct models: one for battery detection and another for device segmentation. The segmentation model generates a mask that identifies the devices in the image, enabling the calculation of their coordinates on the conveyor. These coordinates are subsequently relayed to the robotic system for device removal. For battery detection, we utilize the YOLO model \cite{YOLO2016}, while device segmentation is performed using the U-Net architecture \cite{UNet}.
\subsubsection{Dataset and evaluation method for detection task}
For the detection task, which involves pinpointing the exact location of batteries within the X-ray images, we utilized the YOLO \cite{YOLO2016} model, which is known for its efficiency and accuracy in localizing and classifying objects within images. The key advantage of this system lies in the dual energy configuration, as such, many of the training experiments revolved around finding the best format for utilizing the two energy levels. 


The data collection resulted in a total of $127$ devices being scanned, producing $374$ distinct images. Among these, $270$ contained batteries, while the remaining $104$ images did not. The images with batteries are split into training and validation with $212$ and $58$ images respectively, while those without are split into training and validation with $74$ and $30$ images respectively. The numbers presented here represent unique images. Each unique image has both an HE and a TE representation.

\begin{figure}[!ht]
    \centering
    \includegraphics[width=\linewidth,clip]{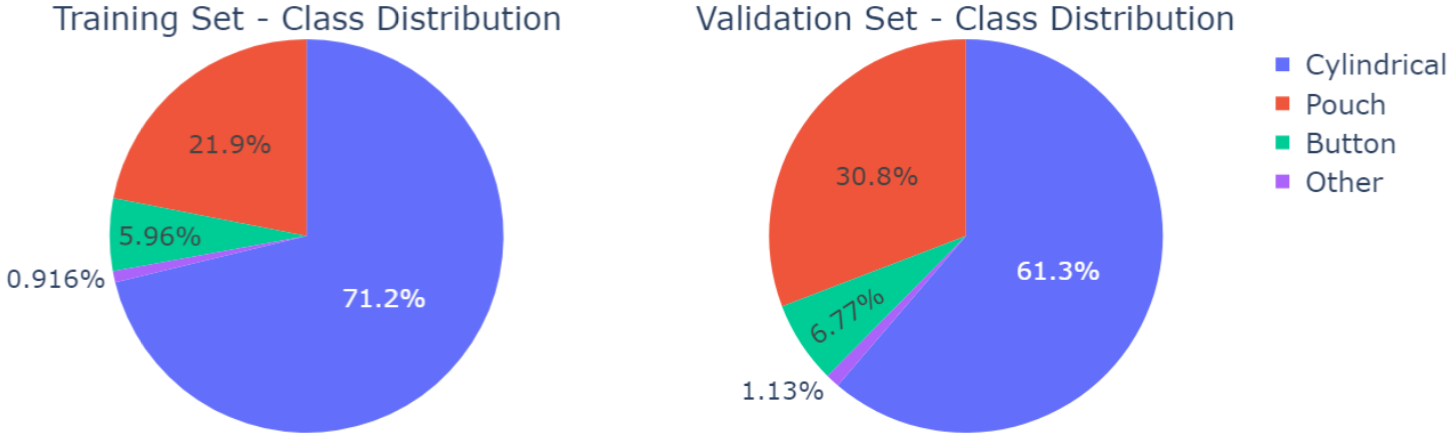}
    \caption{Class distribution in training and validation datasets}
    \label{fig:class_distribution}
\end{figure}

For simplicity reasons the batteries are categorized into four types; Cylindrical, Pouch, Button, and Other, with the last one containing all other lithium and miscellaneous batteries. In Figure \ref{fig:class_distribution}, we demonstrate the distribution of class instances in these sets. Moreover, the dataset was formed in a way to ensure that the devices in the validation set were completely unseen during training. This was achieved by removing images that contain validation set devices from the training set. 

For model evaluation, we use the Recall, Precision, mAP, and mAP (0.5 IoU) as well as a modified version of Recall (which shall be referred to as Modified Recall from this point forward). This is done as in the case of battery detection in waste management streams, pixel-perfect detections are not always required. Batteries in devices are often contained in battery packs, so detecting at least one battery in such a battery cluster is sufficient to count the entire cluster as detected, because this device will be eliminated from the waste stream. To address this we devised a custom algorithm that works as follows: We merge ground truth labels for neighboring batteries for the class of interest. By comparing these merged truth labels with predictions, we determined that if a prediction falls within a ground truth bounding box (i.e. a cluster of batteries), it is counted as a true positive; if no prediction is made for a truth box, it is considered a false negative. By summing all true positives and false negatives across images, we calculate recall in the familiar manner: \({\rm Recall} = {\rm True \, Positives/(\rm True \, Positives + False \, Negatives})\).

 The model was trained for $150$ epochs using a batch size of $10$ and an early stopping patience of $20$, i.e. the training was stopped early if the validation accuracy did not meaningfully improve for $20$ epochs. We also utilized data augmentation including random crops and noise addition, combining the two energy channels. Table I summarizes the results of the performance metrics across each battery category.



\begin{table}[!ht]
    \centering
    \caption{YOLO metrics for the chosen image configuration}
    \rowcolors{2}{white}{gray!20}
    \renewcommand{\arraystretch}{2} %
    \begin{tabular}{m{1.2cm}|m{1cm}cccc}
        \toprule
        \hspace*{-1.2mm}Class    & Modified \hspace*{1.4mm}Recall & Recall & Precision & mAP & mAP (0.5 IoU)\\
        \midrule
        \hspace*{-1.2mm}All   & \hspace{1.9mm}0.920 & 0.864 & 0.854 & 0.659 & 0.899\\
        \hspace*{-1.2mm}Cylindrical  & \hspace{1.9mm}0.988 & 0.963 & 0.931 & 0.687 & 0.979\\
        \hspace*{-1.2mm}Pouch  & \hspace{1.9mm}0.783 & 0.659 & 0.891 & 0.652 & 0.812\\
        \hspace*{-1.2mm}Button  & \hspace{1.9mm}0.909 & 0.833 & 0.686 & 0.554 & 0.808\\
        \hspace*{-1.2mm}Other  & \hspace{1.9mm}1.000 & 1.000 & 0.909 & 0.741 & 0.995\\
        \bottomrule
    \end{tabular}
    \label{tab:yolo_metrics}
\end{table}
From these metrics, we can see that this configuration offers very high values of Modified Recall for every class in our dataset, which is ideal for our use case, as reliable detection of devices is the end goal.

\subsubsection{U-Net Training and results} 
For the device segmentation task, we employed U-Net. U-Net is a CNN primarily designed for biomedical image segmentation but has since been widely used across various fields for image analysis \cite{UNet}. For training U-Net, we used a smaller subset of the dataset used for the detection task, consisting of $41$ images with their corresponding segmentation masks.  
Training for exactly $5$ epochs resulted in almost perfect results, as evidenced by the masks produced for devices that were unseen during training (see Fig. \ref{fig: advanced X-ray imaging}b).

\subsubsection{Combining results from YOLO and U-Net}
\begin{figure}[th]
    \centering
    \includegraphics[width=0.44\textwidth]{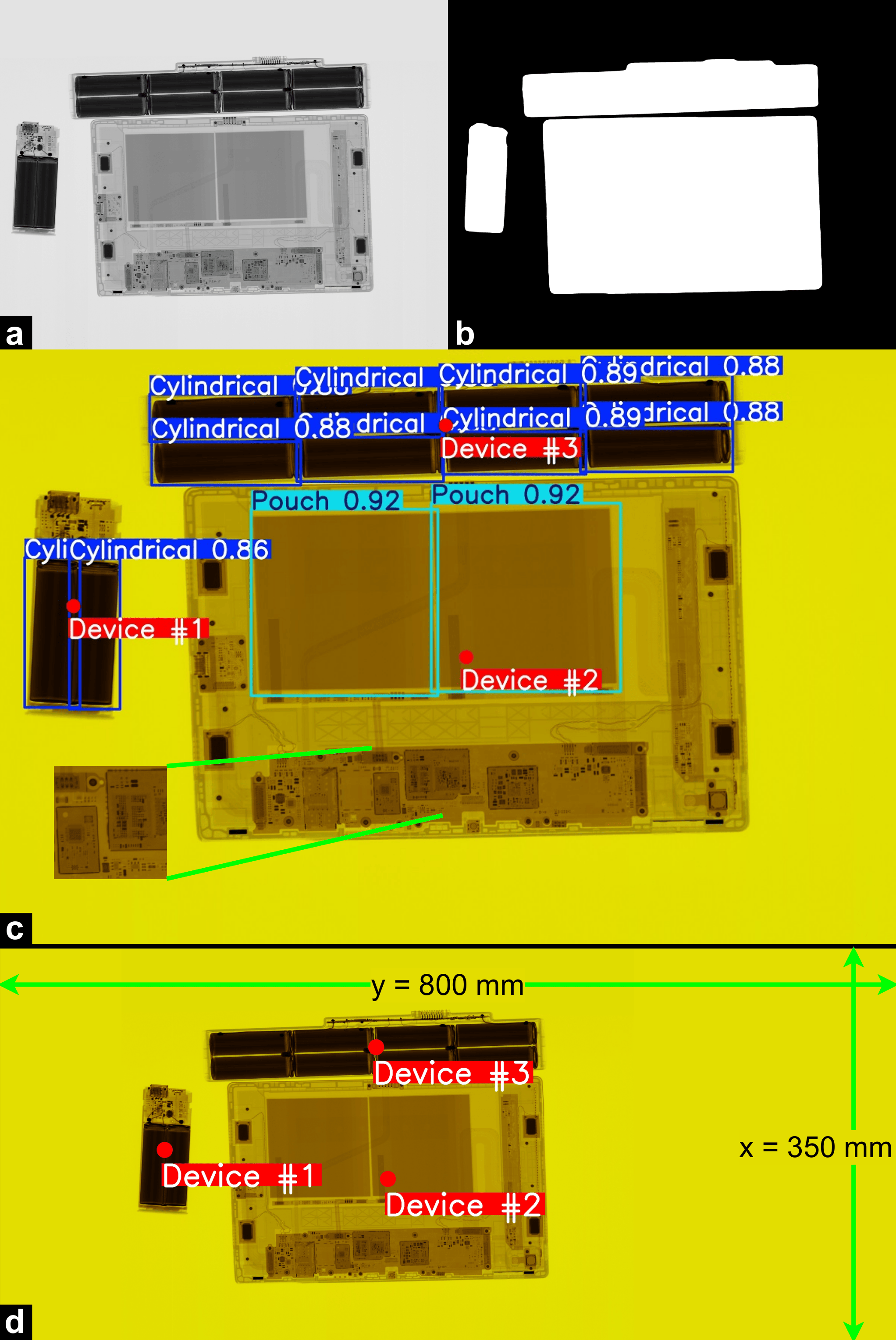}
    \caption{Advanced X-ray imaging, device segmentation and battery detection on a tablet, a power bank and a laptop battery. a) HE X-ray image, b) Device segmentation using U-Net model, c) Battery detection on the reconstructed X-ray image, with annotations of the devices' center, d) Localization for future position estimation of devices with batteries. The width and height of the image correspond to a specific plane on the conveyor belt.}
    \label{fig: advanced X-ray imaging}
\end{figure}

Using a contour detection algorithm, we can convert the device segmentation masks into bounding boxes, that are centered at $(x_{\rm center},\,y_{\rm center})$ and have a specific width and height. Using this information, we can match the battery detections from YOLO to devices by finding the bounding box in which these batteries are contained (see Fig. \ref{fig: advanced X-ray imaging}c). The coordinates of the device's center, as well as the bounding box width and height are then passed on to the robotic system for removal.


\subsection{WEEE Position Estimation}

The conveyor belt's width was selected to match the active area of the X-ray detector, specifically $y = 800$ mm. Given the fixed and known position of the detector, the y-coordinate ($y_{center}$) of each detected battery-containing WEEE in the X-ray image corresponds directly to a spatial coordinate ($y_0$) in the x–y plane, based on a predefined spatial resolution ($mm\_to\_px\_ratio = 0.1$ mm/pixel). Considering the constant conveyor belt velocity, established in Section II-A as $v = 0.35$ m/s, and the detector’s line acquisition rate of $l_r = 3500$ Hz, the detector acquires lines for a duration of $t = 1$ s. Consequently, this results in images with a height of $l_rt = 3500$ pixels, equivalent to a spatial length of $x= vt = 350$ mm (see Fig. 3d). Hence, each x-coordinate ($x_{center}$) in the image is similarly mapped to a corresponding spatial coordinate ($x_0$) in the x–y frame of reference using the same ratio of $0.1$ mm/pixel. A predictive algorithm performs future position estimation by employing the initial spatial coordinates ($x_0$, $y_0$) of all battery-containing devices and their velocity $v$, derived from the output of the detection and segmentation modules along with rotary encoder feedback. To estimate the devices' positions ($x_{t_p}, y_{t_p}$) after a specified time interval $t_p$, the following equation is utilized: $(x_{t_p}, y_{t_p}) = (x_0 + vt_p, y_0)$. The parameter $t_p$ is dependent on the delta robot’s relative position within the defined coordinate reference frame.

\subsection{Robotic Sorting} 

Delta parallel robots are extensively utilized in production lines \cite{DeltaIndustrialApp1} \cite{DeltaIndustrialApp2}, particularly in applications requiring high-speed pick-and-place operations, such as food packaging \cite{DeltaFoodPackaging} and pharmaceutical manufacturing \cite{DeltaPharmaceuticals}. Similarly, in WEEE sorting, a high-speed robotic system can significantly enhance the throughput of the production line. As a result, a custom-designed Delta robot was selected in this work to achieve high performance, providing the additional advantage of flexibility for high-fidelity simulation and safe testing of motion planning algorithms. Specifically, the robot was designed in SolidWorks, allowing precise mass properties and dimensions (Table \ref{table:DeltaRobotParameters}) to be incorporated into a digital twin (see Section \ref{sec:isaac}). For real-time, low-latency performance and synchronized multi-joint control, the EtherCAT protocol \cite{Ethercat} was employed. Each Delta robot arm is actuated by a Leadshine ELM2 servo motor \cite{LeadshineELM2Series}, paired with an EL7-EC400F drive \cite{LeadshineEL7-EC400F} and a speed reduction gearbox (see Table \ref{table:DeltaRobotParameters}). The motors are controlled via PI controllers, with gains tuned using Leadshine’s auto-gain software \cite{LeadshineSWOManual}. High-level position commands are transmitted to the Leadshine drives through the SOEM EtherCAT \cite{GithubOpenEtherCATsocietySOEM} master, running on a Jetson Orin NX. 
A suction gripper has been designed and mounted on the end-effector of the delta robot. The gripper is connected via a tube to a suction pump equipped with a solenoid valve. The solenoid valve control pins are interfaced directly with the Jetson board’s GPIO pins, enabling precise ON/OFF control.



 
\section{Isaac Simulation}\label{sec:isaac}

A simulation framework was developed to facilitate a seamless transition to the real system. NVIDIA Isaac Sim \cite{IsaacSim} was selected due to its support for complex dynamics in closed-kinematic chains, realistic motion and contact modeling, high-fidelity photorealistic virtual environments, and interoperability with robotics and deep learning frameworks.

\subsection{Delta Robot Modeling}

\begin{figure}[!ht]
  \centering
  \includegraphics[width=0.43\textwidth]{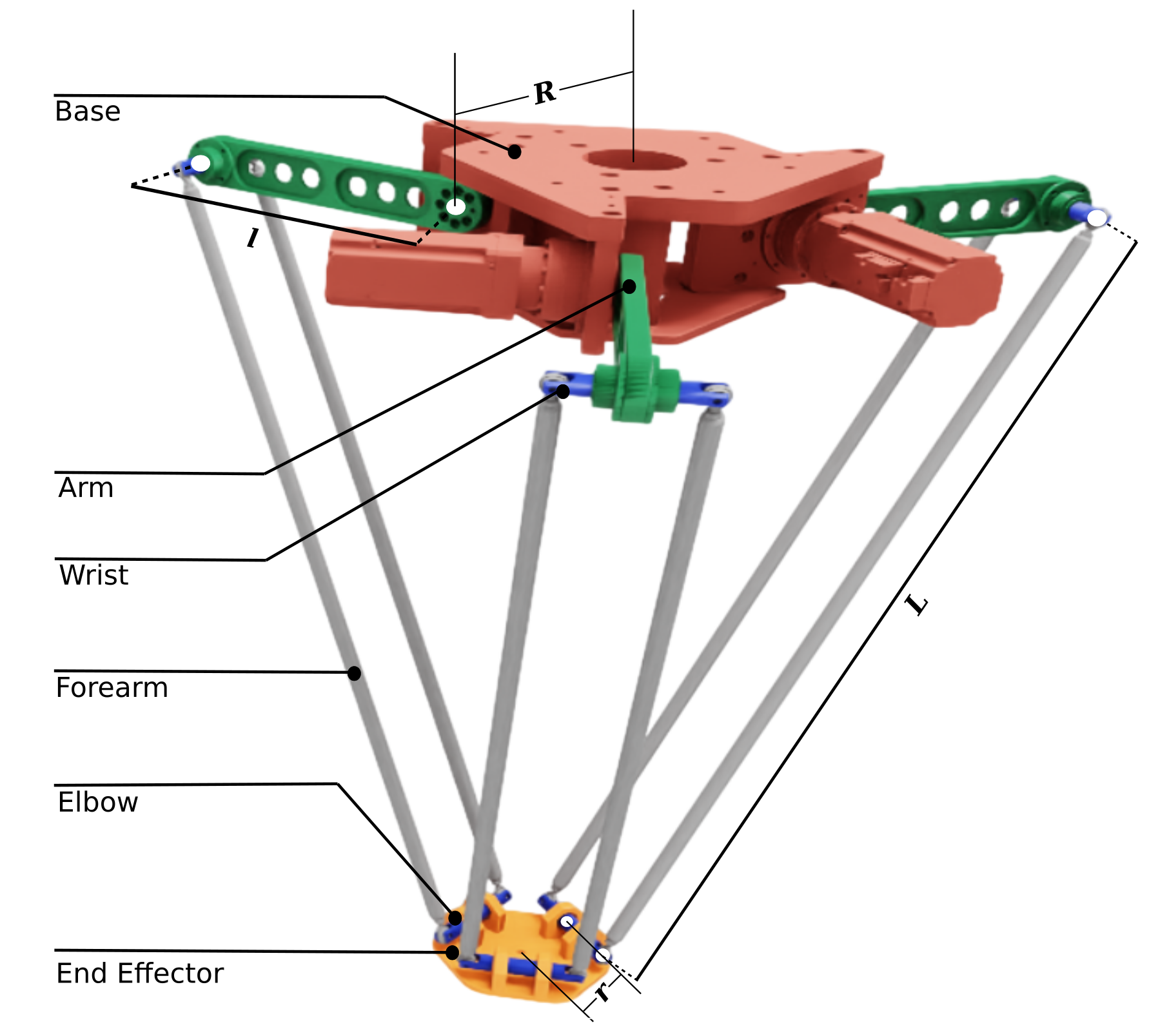} 
  \caption{Detailed model of THL Delta robot in NVIDIA Isaac sim.}
  \label{fig:DeltaModeling}
\end{figure}

Regarding the robot model, STL files from the SolidWorks design for each Delta robot component are imported, along with its mass properties, including mass, inertia matrix, and center of mass. These properties are incorporated into an MJCF \cite{Mujoco} (MuJoCo XML Format) description. Additionally, arm actuators and closed-kinematic constraints are defined using equality constraints. Through the MJCF Importer plugin \cite{IsaacMJCFImporterExtension}, the robot description is seamlessly transferred into the NVIDIA Isaac Sim environment, ensuring precise simulation of its dynamics and interactions. All parameters are summarized in Table \ref{table:DeltaRobotParameters} and Figure \ref{fig:DeltaModeling}.


\begin{table}[ht]
    \centering
    \caption{THL Delta Robot Parameters Modeled in Isaac Sim}
    \resizebox{\linewidth}{!}{%
        \rowcolors{2}{white}{gray!20} 
        \renewcommand{\arraystretch}{1.5} %
        \begin{tabular}{rl}
            \toprule
            \textbf{Parameter} & \textbf{Value} \\
            \midrule
            \makecell[r]{ \newline \\ Total Mass: \{Base, Arm, Forearm, \\ Elbow/Wrist, End Effector\} \\ \qasep } & 
            \makecell[l] {\{$13.74, 0.44, 0.2, 0.05, 0.51$\} } kg  \\
            Base Inertia Matrix & \begin{tabular}{@{}ll}
            I\textsubscript{XX} = $0.0018120034$ kg${\rm m^2}$  \\
            I\textsubscript{YY} = $0.0$ kg${\rm m^2}$ \\
            I\textsubscript{ZZ} = $0.0018120034$ kg${\rm m^2}$ 
            \end{tabular} \\
            Arm Inertia Matrix & \begin{tabular}{@{}ll}
            I\textsubscript{XX} = $0.0001447401$ kg${\rm m^2}$ \\
            I\textsubscript{YY} = $0.015154084$ kg${\rm m^2}$ \\
            I\textsubscript{ZZ} = $0.0152228384$ kg${\rm m^2}$ 
            \end{tabular} \\
            Forearm Inertia Matrix & \begin{tabular}{@{}ll}
            I\textsubscript{XX} = $0.016870159$ kg${\rm m^2}$ \\
            I\textsubscript{YY} = $0.0000088926$ kg${\rm m^2}$ \\
            I\textsubscript{ZZ} = $0.0168698215$ kg${\rm m^2}$ 
            \end{tabular} \\
            Wrist/Elbow Inertia Matrix & \begin{tabular}{@{}ll}
            I\textsubscript{XX} = $0.0000365556$ kg${\rm m^2}$ \\
            I\textsubscript{YY} = $0.0000019483$ kg${\rm m^2}$ \\
            I\textsubscript{ZZ} = $0.0000364106$ kg${\rm m^2}$ 
            \end{tabular} \\
            End Effector Inertia Matrix & \begin{tabular}{@{}ll}
            I\textsubscript{XX} = $0.0000679006$ kg${\rm m^2}$ \\
            I\textsubscript{YY} = $0.0$ kg${\rm m^2}$ \\
            I\textsubscript{ZZ} = $0.0000679006$ kg${\rm m^2}$ 
            \end{tabular} \\
            Arm Reduction Ratio & 25:1 \\
            Base Radius (\textit{R}) \& End Effector Radius (\textit{r}) & $150.0$ mm \& $54.0$ mm\\
            Arm Length (\textit{l}) \& Forearm Length (\textit{L}) & $260.0$ mm \& $820.0$ mm\\
            \bottomrule
        \end{tabular}
    }
    \label{table:DeltaRobotParameters}
\end{table}

\subsection{Delta Robot Kinematics and Trajectory Planning}

The kinematic model of the delta robot, depicted in Figure \ref{fig:DeltaModeling}, defines the relationship between the actuated joint angles $\theta_1, \theta_2 $ and $\theta_3$ and the position of the end-effector (EEF), represented by $P = \left[x_p, y_p, z_p \right]$ \cite{DeltaKinematics}. Due to the symmetric arrangement of the robot’s arms around its base coordinate system, the kinematic computations can be simplified by defining the configuration angles as $\gamma_i = [0^o, 120^o, 240^o]$, where $\gamma_i$ indicates the relative angular offset between the base coordinate frame and arm link i. Equations (\ref{Eq:delta_fk}) and (\ref{Eq:delta_ik}) describe the forward and inverse kinematic models of the delta robot, respectively, for $i = [1,2,3]$, utilizing the corresponding parameters presented in Table \ref{table:DeltaRobotParameters}.

\begin{gather} \label{Eq:delta_fk} 
    x_{p}^{2}+y_{p}^{2}+z_{p}^{2}+a_{i} x_{p}+b_{i} y_{p}+c_{i} z_{p}+d_{i} =0
\end{gather}
where:
\begin{gather*}
    a_{i} =  2 ( R - r + l \cos \theta_{i} ) \cos \gamma_{i} \\ 
    b_{i} = 2 ( R - r + l \cos \theta_{i} ) \sin \gamma_{i} \\ 
    c_{i} = 2 l \sin \theta_{i}  \\ 
    d_{i} = l^{2}-L^{2}+(R - r)^{2}+ 2 (R-r) l \cos \theta_{i} 
\end{gather*}

\begin{gather} \label{Eq:delta_ik}
    \theta_{i} = 2 \tan^{-1} \left ( \dfrac{-B_i \pm \sqrt{B^2_i - C_i + A_i}}{C_i - A_i}\right)
\end{gather}
where: 
\begin{gather*}
    A_i = 2 l x_p \cos \gamma_i + 2 L y_p \sin \gamma_i + 2 (R - r) l \\
    B_i = 2 l z_p \\
    C_i = l^2 - L^2 + (R - r)^2 + 2 (R -r) cos \gamma_i x_p +  \\ 
    2 (R - r) \sin \gamma_i y_p + x^2_p + y^2_p + z^2_p 
\end{gather*}

\begin{figure}[!ht]
  \centering
  \includegraphics[width=0.4\textwidth]{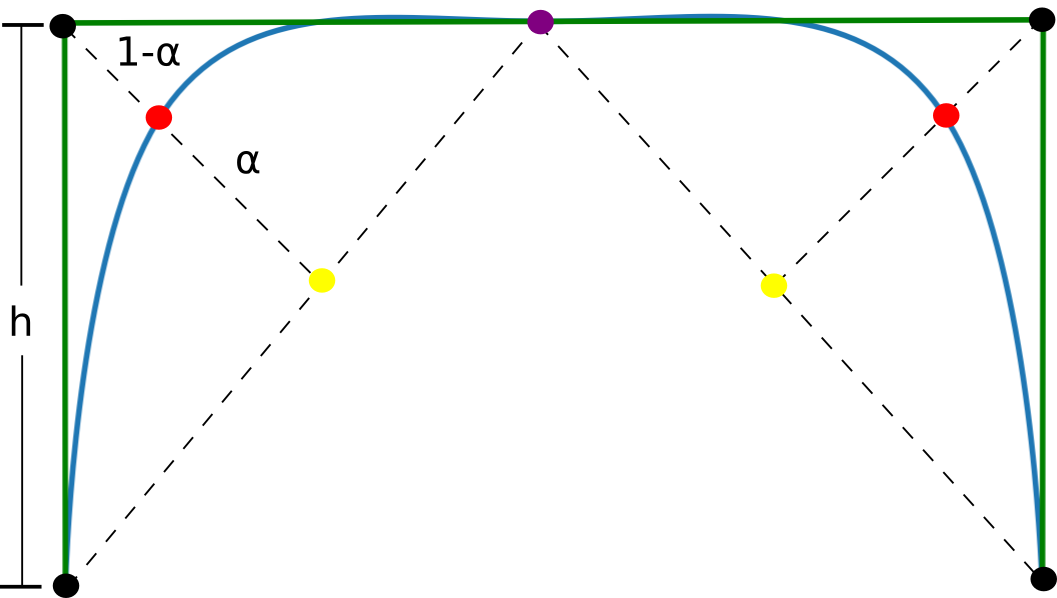} 
  \caption{The green trajectory represents a 3D Pi-shaped path, where the intermediate points maintain a constant offset $h$. The purple point is determined as the median of the two intermediate points. A curvature factor $\alpha$ is applied to adjust the intermediate points accordingly. Using this median point, the minimum acceptable values are computed and highlighted in yellow. This approach ensures that the generated trajectory (blue) remains constrained within the boundaries defined by the Pi-shaped path.}
  \label{fig:TrajectoryExplanation}
\end{figure}

For the pick-and-place tasks, a four-point 3D Pi-shaped trajectory (see Fig. \ref{fig:TrajectoryExplanation}) is used, defined in Cartesian coordinates relative to the Delta robot’s end effector coordinate system \cite{DeltaTrajectory}. Given total time $t_n$ and initial time \(t_0 = 0\), the trajectory is generated using cubic splines (\(p = 3\)), chosen for their smooth velocity and continuous acceleration. This avoids mechanical stress, excessive torque demands, and vibrations, ensuring a smooth motion profile \cite{DeltaTrajectoryVibration}. The $4$-point defined trajectory interpolation that uses \(n=3\) splines can be described as
\begin{gather} \label{eq:delta_trajectory}
    s ( t )=\{q_{k} ( t ), \ t \in[ t_{k}, \, t_{k+1} ], \ k=0, \ldots, n-1 \}
\end{gather}
where:
\begin{gather}
    q_{k} ( t )=a_{k 0}+a_{k 1} ( t-t_{k} )+a_{k 2} ( t-t_{k} )^{2}+ \nonumber \\
    a_{k 3} ( t-t_{k} )^{3} 
\end{gather}

Equations (\ref{eq:initial_condition_1}), (\ref{eq:initial_condition_2}), (\ref{eq:initial_condition_3}) and (\ref{eq:initial_condition_4}) represent the initial conditions for spline generation and Equation system (\ref{eq:spline_coeffs}) computes the coefficients $a_{k,i}$ for \(i = 0, \cdots, n-1\). Intermediate velocity points can be calculated using the method presented in \cite{DeltaVelocity}. The trajectory described in Equation (\ref{eq:delta_trajectory}) is defined in the end-effector Cartesian space. By applying inverse kinematics (Eq. \ref{Eq:delta_ik}), it is transformed into the joint space of the Delta robot.
\begin{gather}
    q_{k} ( t_{k} )=\theta_{k}, \quad q_{k} ( t_{k+1} )=\theta_{k+1}, \ k=0, \ldots, n-1 \label{eq:initial_condition_1} \\ 
    \dot{q}_{k} ( t_{k+1} )=\dot{q}_{k+1} ( t_{k+1} ), \ {{k=0, \, \ldots, n-2}} \label{eq:initial_condition_2} \\ 
    \dot{q}_{k} ( t_{k+1} )=\dot{q}_{k+1} ( t_{k+1} ), \ {{k=0, \, \ldots, n-2}} \label{eq:initial_condition_3} \\ 
    \dot{q}_{0} ( t_{0} )=v_{0}=0, \quad\dot{q}_{n-1} ( t_{n} )=v_{n}=0 \label{eq:initial_condition_4}
\end{gather}

\begin{gather} \label{eq:spline_coeffs}
    \left\{\begin{aligned} {{a_{k, 0}}} & {{} {{} {} {{}=q_{k}}}} \\ {{a_{k, 1}}} & {{} {{} {{}=v_{k}}}} \\ {{a_{k, 2}}} & {{} {{} {} {} {}=\frac{1} {T_{k}} \left[ \frac{3 ( q_{k+1}-q_{k} )} {T_{k}}-2 v_{k}-v_{k+1} \right]}} \\ {{a_{k, 3}}} & {{} {} {{} {} {} {}=\frac{1} {T_{k}^{2}} \left[ \frac{2 ( q_{k}-q_{k+1} )} {T_{k}}+v_{k}+v_{k+1} \right]}} \\ \end{aligned} \right.
\end{gather}

\subsection{Simulation Environment}

A simulation environment for testing the pick-and-place pipeline was developed in Isaac Sim. It features the THL Delta robot equipped with a suction gripper utilizing the Surface Gripper Extension \cite{IsaacSim}, a constant-speed conveyor belt, and a bin where the battery-containing items are discarded (Figure \ref{fig: Grinner setup}). Smartphone meshes are periodically spawned at the conveyor, where a random subset of items is designated as battery-containing. This information is transmitted to the Delta robot to initiate the pick-and-place process. The task consists of two phases: (1) Trajectory generation and tracking to precisely control the end-effector, enabling it to catch the target object, (2) Item rejection, where the robot places the item into the designated bin. This process repeats continuously throughout the simulation, ensuring an efficient and systematic evaluation of the pick-and-place pipeline. Trajectory and grasp commands are executed within the same control loop as the real robot.


\section{Sorting Procedure} \label{sec:sorting_procedure}

\subsection{Procedure Overview}

\begin{figure}[ht]
    \centering
    \includegraphics[width=0.49\textwidth]{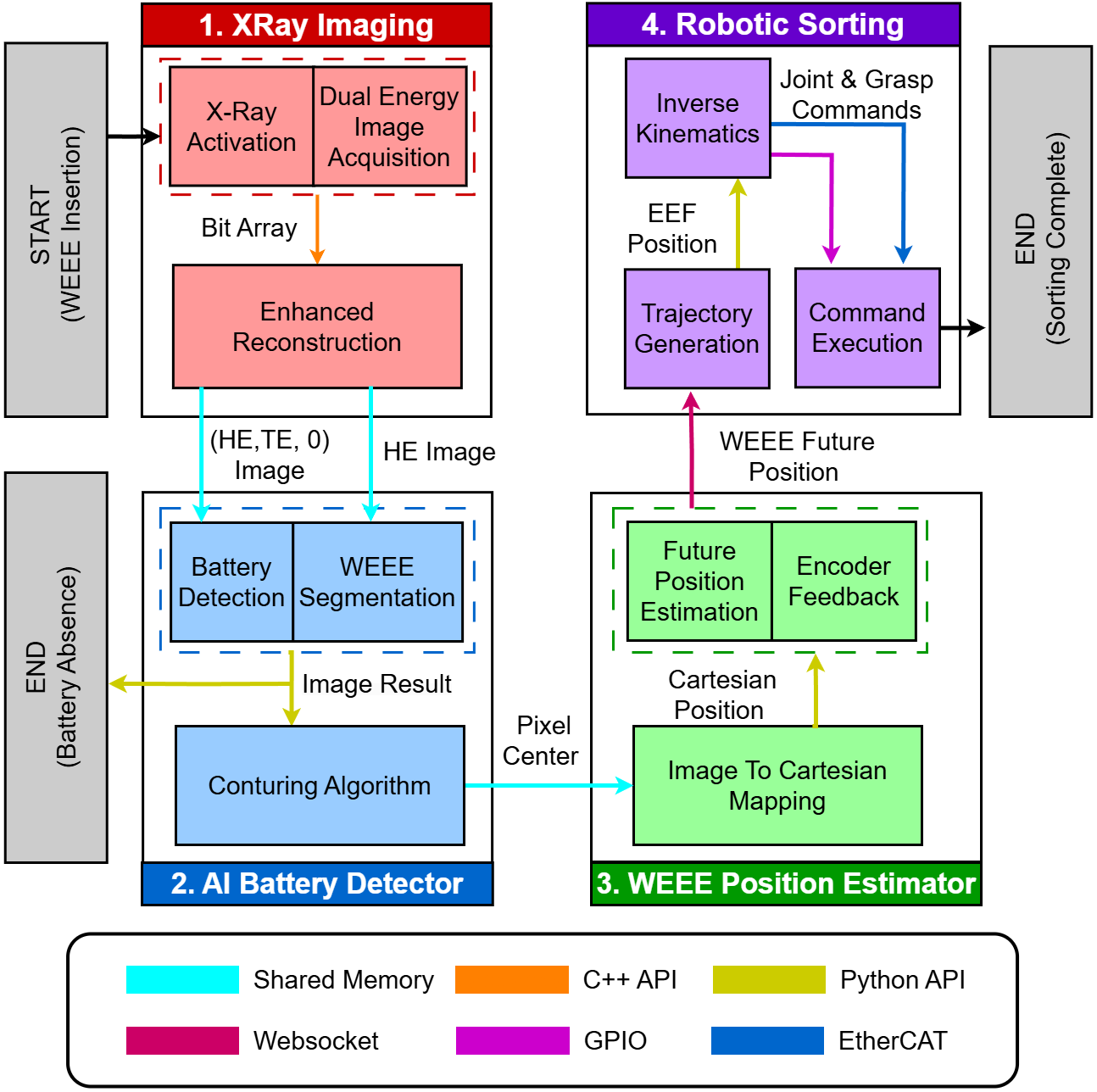}
    \caption{Software integration and communication flow in the presented sorting procedure, highlighting key processing stages and the interaction of various modules through diverse communication protocols.}
    \label{fig:the proposed sorting procedure}
\end{figure}

Our system has been designed to execute the sorting process depicted in Figure \ref{fig:the proposed sorting procedure}, comprising four integrated modules: (1) X-ray Imaging, (2) AI-Based Battery Detection, (3) WEEE Position Estimation, and (4) Robotic Sorting. The procedure begins in the X-ray imaging module, where WEEE items are placed within the imaging chamber, and high-resolution X-ray images are reconstructed. Subsequently, in the AI-based battery detection module, YOLO and U-Net neural networks perform battery detection and device segmentation, respectively. These outputs are integrated with a contour detection algorithm to accurately determine the geometric center of battery-containing WEEE items requiring removal from the conveyor belt. If no batteries are detected, the sorting process concludes at this stage. Otherwise, the WEEE position estimation module converts the identified image coordinates into spatial Cartesian coordinates. Additionally, this module utilizes encoder feedback to predict the future positions of detected battery-containing devices. Finally, in the robotic sorting module, a delta robot uses these estimated spatial coordinates, inverse kinematics calculations, and trajectory generation algorithms to accurately and efficiently remove the identified items from the conveyor belt.

\subsection{Software Architecture}

The software modules responsible for X-ray image acquisition, AI-based detection, and WEEE position estimation are executed on a computer system equipped with a $2.6 GHz$ Intel i7 CPU and $16 GB$ of RAM, running the Windows 11 operating system. Upon triggering the X-ray detector, custom-developed software acquires dual-energy (TE and HE) images represented as two-dimensional arrays, each dimension corresponding to the height and width of the captured X-ray image. These raw data arrays undergo preprocessing to generate reconstructed high-energy (HE) images, which are subsequently input to the AI detection module. This module loads pre-trained neural network models designed specifically for battery detection and WEEE segmentation tasks, executing inference to identify relevant objects. Subsequently, a contour detection algorithm, leveraging functionalities from the OpenCV library, calculates the geometric center of each detected object within the image.

The X-ray images and corresponding detected object coordinates are communicated between software modules via shared memory streams. Following this, the detected image coordinates of battery-containing WEEE items are mapped into Cartesian coordinates, initiating the position estimation procedure. A WebSocket server transmits these estimated future positions, along with designated pick-and-place timings. A WebSocket client receives these instructions and generates the appropriate motion trajectories. The end-effector (EEF) Cartesian coordinates are converted into joint-space commands via inverse kinematic computations. These joint commands are transmitted through an SOEM EtherCAT master to the EtherCAT slave motor controllers for execution. At the conclusion of the pick-and-place sequence, GPIO pins controlling the suction gripper are toggled to enable or disable suction for item manipulation. Additionally, from a simulation standpoint, an equivalent WebSocket client receives and executes the identical commands to validate and optimize the generated robotic trajectories.


\section{Experimental Results}

\begin{figure}[!ht]
  \centering
  \includegraphics[width=0.44\textwidth]{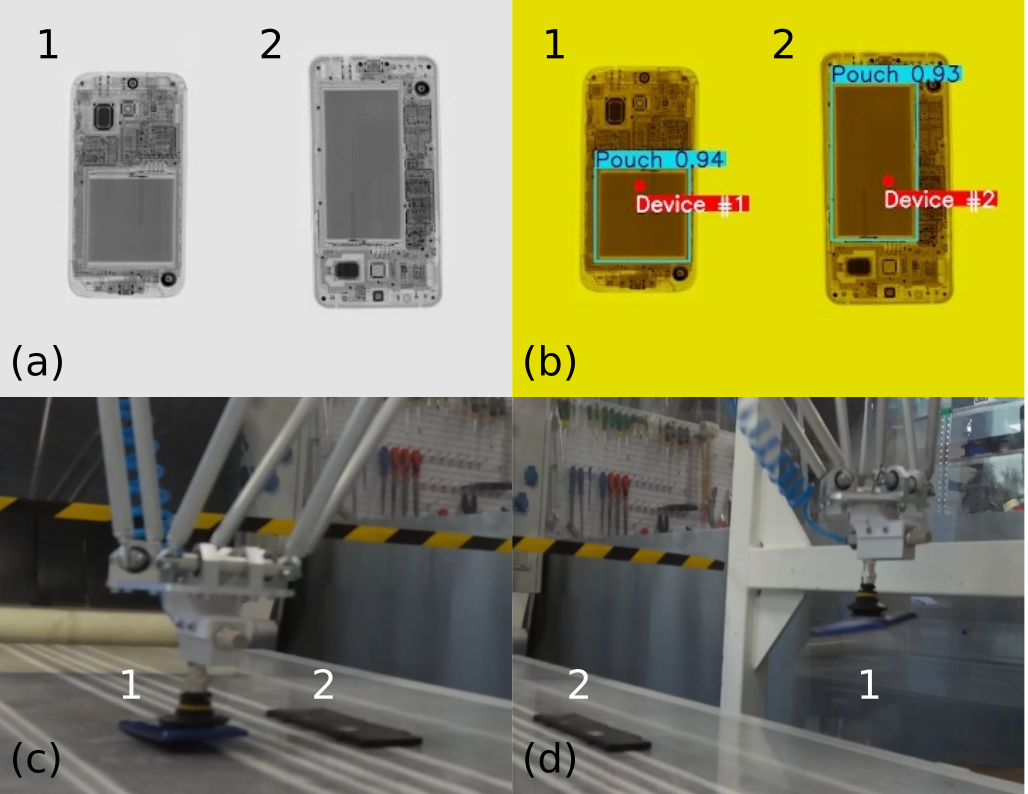} 
  \caption{Cellphones are passed through the conveyor belt for battery detection and sorting. (a) X-ray HE image (b) Battery detection and device center annotation (c) Pick trajectory and grasping of detected item (d) Place trajectory execution}
  \label{fig:PipelineExperiment}
\end{figure}

To evaluate the effectiveness of our proposed approach, a total of 120 WEEE items were inspected following the procedure detailed in Section IV, utilizing the integrated systems deployed at the THL facility (illustrated in Figure \ref{fig: Grinner setup}). For the validation process, the dataset consisted of $70\%$ battery-containing and $30\%$ non-battery-containing devices. All WEEE items featured flat surfaces and were manually loaded into the X-ray imaging chamber in batches. The conveyor belt was configured to operate at a consistent speed of $0.35~{\rm m/s}$, and the delta robot was placed at a distance of 2 meters downstream from the X-ray chamber, aligned with the conveyor belt’s direction of motion. Figure \ref{fig:PipelineExperiment} provides a representative snapshot of the sorting operation in progress.

\begin{figure}[!ht]
  \centering
  \includegraphics[width=0.47\textwidth]{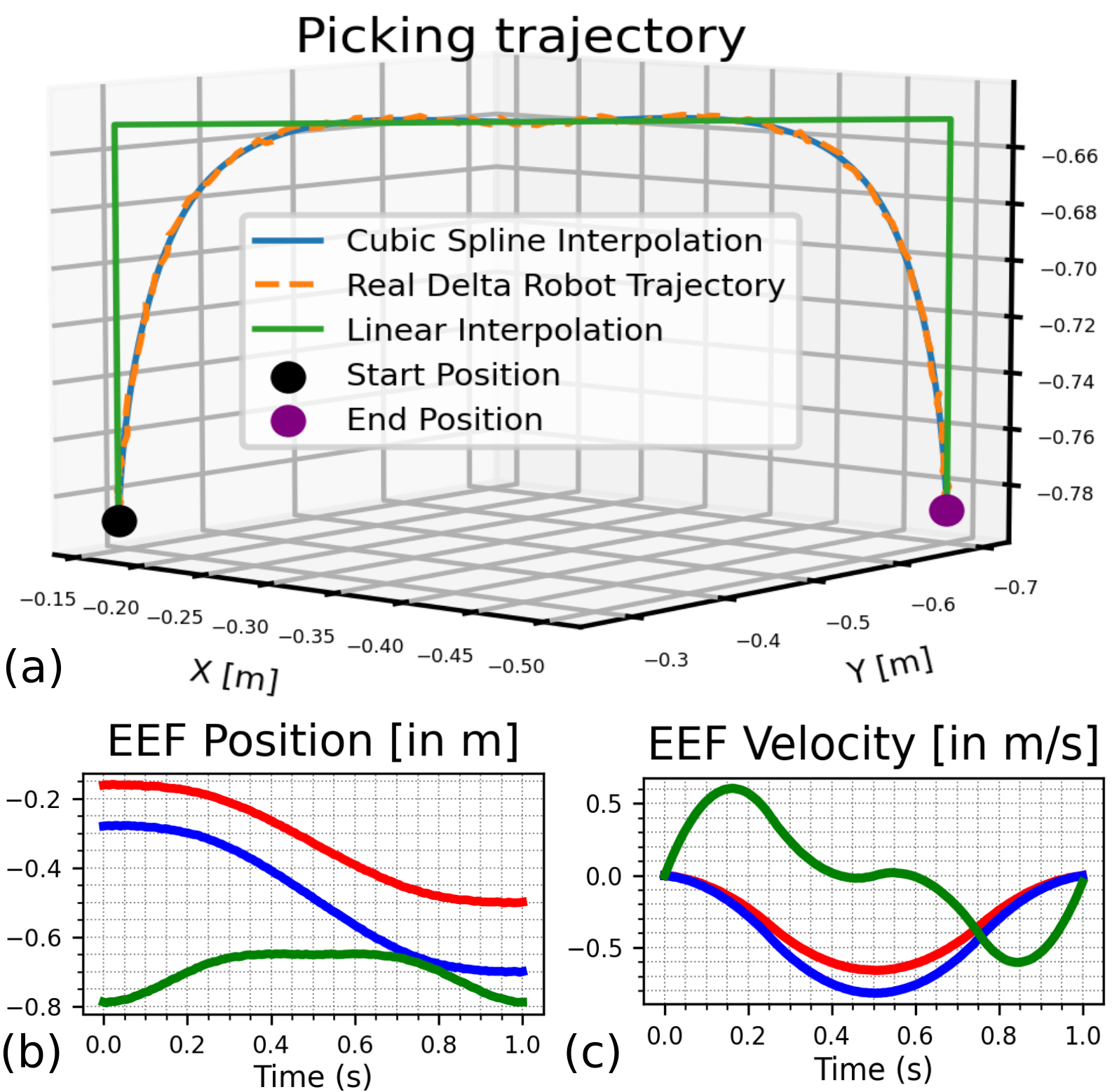} 
  \caption{Executed picking trajectory for battery contained cellphone sorting, with total duration $t_n - t_0 = 1 \ sec$ and curvature factor $\alpha = 0.77$. (a) 3D trajectory comparison alongside the actual Delta robot path (b) EEF position. (c) EEF velocity profile. }
  \label{fig:DeltaTrajectoryPlots}
\end{figure}

The delta robot successfully sorted all detected battery-containing WEEE devices. The executed trajectory, depicted in Figure \ref{fig:DeltaTrajectoryPlots}a, illustrates minimal discrepancies between the actual (orange) and reference (blue) trajectories. The slight variations between aforementioned trajectories confirm that the developed simulation environment can effectively serve as a digital twin of the entire sorting pipeline, enabling testing and optimization without disrupting physical operations. Furthermore, a comparison with the abrupt changes in the linear interpolation (green) shows the smoothness of proposed trajectory generation algorithm. Both the end-effector (EEF) position and velocity profiles, as illustrated in Fig. \ref{fig:DeltaTrajectoryPlots}b \& \ref{fig:DeltaTrajectoryPlots}c exhibit smooth transitions, ensuring gradual acceleration \& deceleration. This performance confirms stable operation and mechanical reliability of the sorting system.


\section{Conclusion}

In this paper, we present a fully automated framework designed for the precise detection and sorting of batteries embedded within flat-surfaced Waste Electrical and Electronic Equipment (WEEE). The proposed system integrates high-resolution X-ray imaging with AI-based object detection and segmentation algorithms, enabling accurate identification of battery-containing devices. Effective and efficient sorting is facilitated by an advanced position estimation module combined with a custom-engineered delta robot manipulator, capable of executing precise pick-and-place operations on a continuously moving conveyor belt. The coordinated interaction among these subsystems ensures high-speed, accurate, and high-throughput performance, highlighting substantial potential for improving current recycling processes. Future research directions include expanding the system's functionality to accommodate WEEE items with varying geometries and dimensions, thereby enhancing adaptability and enabling validation across diverse recycling scenarios. Additionally, continuous refinement of detection algorithms is expected to further improve identification accuracy in overlapping WEEE, while the optimization of robotic motion trajectories will be explored to maximize operational throughput.

\end{document}